\documentclass[runningheads]{llncs}

\usepackage[paperheight=233mm,paperwidth=152mm,%
  textwidth=12.2cm,  % llncs has 12.2cm
  textheight=19.3cm, % llncs has 19.3cm
  hratio=1:1,        % horizontally centered
  vratio=1:1,        % not vertically centered
  ]{geometry}
  
\usepackage{bm} 
\newcommand\Bb{\bm{b}}%
\newcommand\Bc{\bm{c}}%
\newcommand\Bi{\bm{i}}%
\newcommand\Bo{\bm{o}}%
\newcommand\Bf{\bm{f}}%
\newcommand\Bx{\bm{x}}%
\newcommand\By{\bm{y}}%
\newcommand\Bz{\bm{z}}%
\newcommand\BU{\bm{U}}%
\newcommand\BW{\bm{W}}%
\usepackage{amsfonts}   % blackboard math symbols
\usepackage[hidelinks]{hyperref}
\hypersetup{
  colorlinks    = true, 
  urlcolor      = black,
  linkcolor     = blue, 
  citecolor     = blue }
\usepackage[misc]{ifsym}
\newcommand{\Arrrepeatthanks}{\textsuperscript{\thefootnote}} 
\newcommand{\changeurlcolor}[1]{\hypersetup{urlcolor=#1}} 
\makeatletter
\def\blfootnote{\xdef\@thefnmark{}\@footnotetext}
\makeatother
\usepackage{amsmath}
\usepackage{amssymb}
\usepackage{xcolor}
\usepackage{graphicx}
\usepackage{array}
\usepackage{multirow}
\usepackage{makecell}
\usepackage{xspace}
\usepackage{cleveref}
\usepackage[nolist]{acronym}
\usepackage{listings}
\usepackage{lipsum}
\usepackage{booktabs}
\usepackage{footnote}
\usepackage{bm} 
\usepackage{placeins}
\usepackage{enumitem}
\usepackage{emptypage}
\usepackage{imakeidx}

\title{Explaining and Interpreting LSTMs}
\author{Leila Arras\thanks{L. Arras and J. Arjona-Medina contributed equally to this work.}$^1$, Jos{\'e} Arjona-Medina\Arrrepeatthanks$^2$, Michael Widrich$^2$, Gr{\'e}goire Montavon$^3$, Michael Gillhofer$^2$, Klaus-Robert M{\"u}ller$^{3,4,5}$, Sepp Hochreiter$^2$, \\ and Wojciech Samek$^{1{\textrm{(\Letter)}}}$}
\institute{$^1$ Fraunhofer Heinrich Hertz Institute, 10587 Berlin, Germany\\
\texttt{\{leila.arras,wojciech.samek\}@hhi.fraunhofer.de}\\
$^2$Johannes Kepler University Linz, 4040 Linz, Austria\\
\texttt{\{arjona,widrich,gillhofer,hochreit\}@ml.jku.at}\\
$^3$Technische Universit{\"a}t Berlin, 10587 Berlin, Germany\\
\texttt{\{gregoire.montavon,klaus-robert.mueller\}@tu-berlin.de}\\
$^4$Korea University, Anam-dong, Seongbuk-gu, Seoul 02841, Korea\\
$^5$Max Planck Institute for Informatics, 66123 Saarbr{\"u}cken, Germany
}

\begin{document}

\authorrunning{Arras et al.}
\titlerunning{Explaining and Interpreting LSTMs}

\maketitle

\begin{abstract}
While neural networks have acted as a strong unifying force in the design of modern AI systems, the neural network architectures themselves remain highly heterogeneous due to the variety of tasks to be solved. In this chapter, we explore how to adapt the Layer-wise Relevance Propagation (LRP) technique used for explaining the predictions of feed-forward networks to the LSTM architecture used for sequential data modeling and forecasting. The special accumulators and gated interactions present in the LSTM require both a new propagation scheme and an extension of the underlying theoretical framework to deliver faithful explanations.
\end{abstract}

\section{Introduction}

\blfootnote{\scriptsize The final authenticated publication is available online at {\changeurlcolor{blue}\url{https://doi.org/10.1007/978-3-030-28954-6_11}}. In: W. Samek et al. (Eds.) Explainable AI: Interpreting, Explaining and Visualizing Deep Learning. Lecture Notes in Computer Science, vol 11700, pp. 211-238. Springer, Cham (2019)}

In practical applications, building high-performing AI systems is not always the sole objective, and interpretability may also be an important issue \cite{Arr:EU-GDPR}.

Most of the recent research on interpretable AI has focused on feedforward neural networks, especially the deep rectifier networks and variants used for image recognition \cite{Arr:Simonyan:ICLR2014,Arr:Zeiler:ECCV2014}. Layer-wise relevance propagation (LRP) \cite{Arr:Bach:15,Arr:montavon2019overview} was shown in this setting to provide for state-of-the-art models such as VGG-16, explanations that are both informative and fast to compute, and that could be embedded in the framework of deep Taylor decomposition \cite{Arr:Montavon:PR2017}.

However, in the presence of sequential data, one may need to incorporate temporal structure in the neural network model, e.g. to make forecasts about future time steps. In this setting it is key to be able to learn the underlying {\em dynamical system}, e.g.\ with a recurrent neural network, so that it can then be simulated forward.
Learning dynamical systems with long-term dependencies using recurrent neural networks presents a number of challenges. 
The backpropagation through time learning signal tends to either blow up or vanish \cite{Arr:Hochreiter:91,Arr:Bengio:94}.
To reduce this difficulty, special neural network architectures have been proposed, in particular, the Long Short-Term Memory (LSTM) \cite{Arr:Hochreiter:91,Arr:Hochreiter:95,Arr:Hochreiter:97}, which makes use of special accumulators and gating functions.

The multiple architectural changes and the unique nature of the sequential prediction task make a direct application of the LRP-type explanation technique non-straightforward. To be able to deliver accurate explanations, one needs to carefully inspect the structure of the LSTM blocks forming the model and their interaction.

In this chapter, we explore multiple dimensions of the interface between the LRP technique and the LSTM. First, we analyze how the LRP propagation mechanism can be adapted to accumulators and gated interactions in the LSTM. Our new propagation scheme is embedded in the deep Taylor decomposition framework \cite{Arr:Montavon:PR2017}, and validated empirically on sentiment analysis and on a toy numeric task. Further, we investigate how modifications of the LSTM architecture, in particular, on the cell input activation, the forget and output gates and on the network connections, make explanations more straightforward, and we apply these changes in a reinforcement learning showcase.

The present chapter elaborates on our previous work \cite{Arr:Arjona-Medina:18,Arr:Arras:19}.

\section{Background}

\subsection{Long Short-Term Memory (LSTM)}
\label{Arr:sec:LSTM}

\index{Neural Networks!LSTM}
\index{LSTM}

Recently, {\em Long Short-Term Memory} (LSTM; \cite{Arr:Hochreiter:91,Arr:Hochreiter:95,Arr:Hochreiter:97})
networks have emerged as the best-performing technique in speech and language processing.
LSTM networks have been overwhelmingly successful in different speech and language applications,
including handwriting recognition \cite{Arr:Graves:09}, 
generation of writings \cite{Arr:Graves:14arxivb},
language modeling and identification \cite{Arr:Gonzalez-Dominguez:14,Arr:Zaremba:14arxiva},
automatic language translation \cite{Arr:Sutskever:14nips}, 
speech recognition \cite{Arr:Sak:14,Arr:Geiger:14},
analysis of audio data \cite{Arr:Marchi:14}, analysis, annotation, and
description of video data \cite{Arr:Donahue:14,Arr:Venugopalan:14,Arr:Srivastava:15}. 
LSTM has facilitated recent benchmark records in TIMIT phoneme recognition,
optical character recognition, text-to-speech synthesis,
language identification, large vocabulary speech recognition,
English-to-French translation, audio onset detection, social signal classification,
image caption generation, video-to-text description, end-to-end speech recognition,
and semantic representations.

The key idea of LSTM is the use of memory cells that allow for constant error flow
during training. Thereby, LSTM avoids the {\em vanishing gradient problem}, that is,
the phenomenon that training errors are decaying when they are back-propagated through time
\cite{Arr:Hochreiter:91,Arr:Hochreiter:00}.
The vanishing gradient problem severely impedes {\em credit assignment} in recurrent neural
networks, i.e.\ the correct identification of relevant events whose effects are not
immediate, but observed with possibly long delays.
LSTM, by its constant error flow, avoids vanishing gradients and, hence, allows for
{\em uniform credit assignment}, i.e.\ all input signals obtain a similar error signal.
Other recurrent neural networks are not able to assign the same credit 
to all input signals and therefore, 
are very limited concerning the solutions they will
find. Uniform credit assignment enables LSTM networks to excel in speech and
language tasks: if a sentence is analyzed, then the first word can be as important as
the last word. Via uniform credit assignment, LSTM networks regard all words of a sentence equally.
Uniform credit assignment enables to consider all input information
at each phase of learning, no matter where it is located in the input
sequence. Therefore, uniform credit assignment reveals many more
solutions to the learning algorithm, which would otherwise remain hidden.

\paragraph{LSTM in a Nutshell.}
% ----------------------------------------------------------------------------------------------------------------------------------------------
\begin{figure}[!b]
\centering
\includegraphics[angle=0,width=1.0\textwidth]{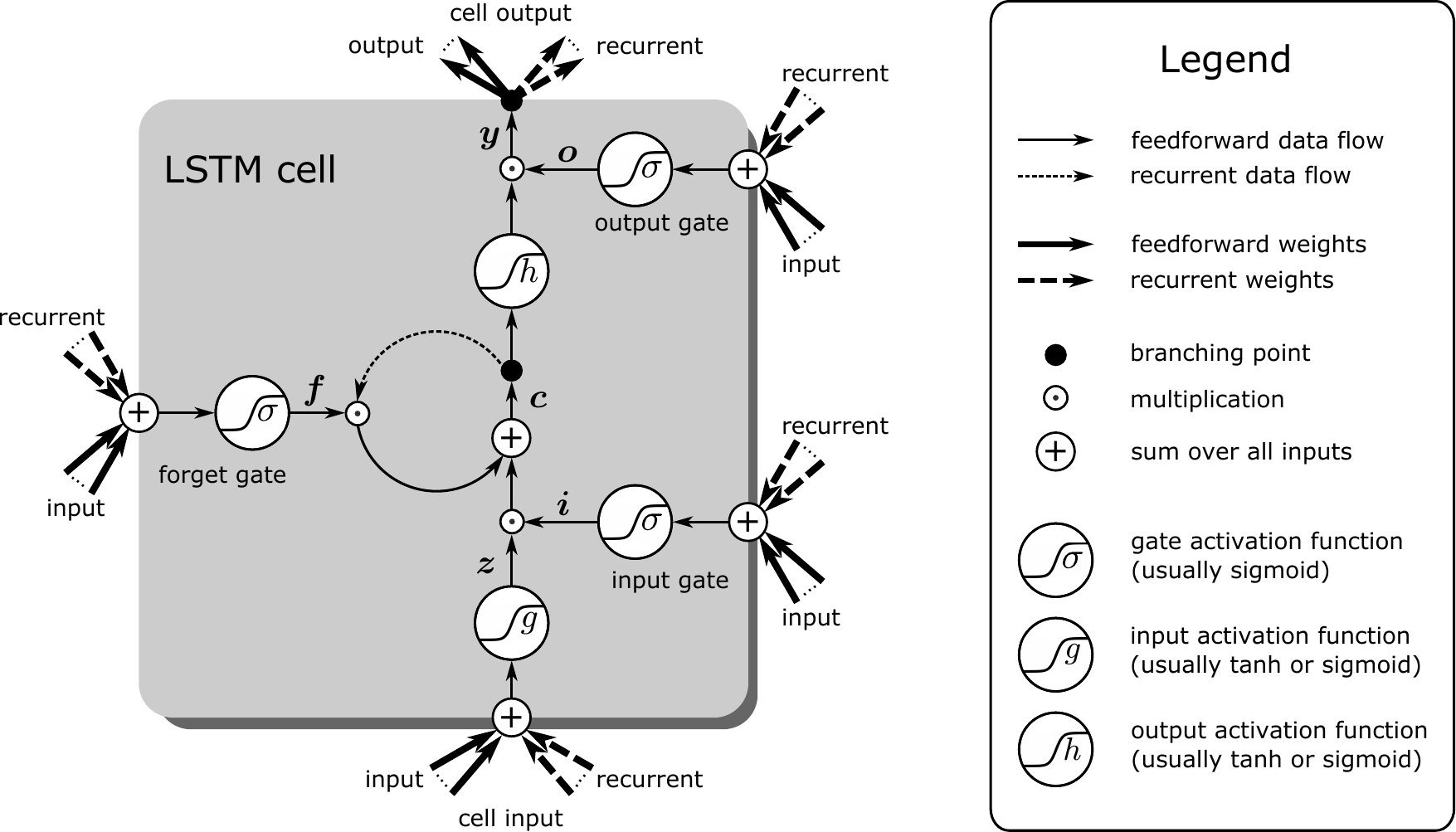}
\caption{LSTM memory cell without peepholes. 
$\Bz$ is the vector of cell input
activations, $\Bi$ is the vector of input gate
activations,  $\Bf$ is the vector of forget gate
activations,  $\Bc$ is the vector of memory cell states,
$\Bo$ is the vector of output gate
activations, and $\By$ is the vector of cell output 
activations. The activation functions are $g$ for the cell input, $h$ for the cell
state, and $\sigma$ for the gates. Data flow is either ``feed-forward''
without delay or ``recurrent'' with a one-step delay.
``Input'' connections are from the
external input to the LSTM network, while ``recurrent'' connections take inputs
from other memory cell outputs $\By$ in the LSTM network with a delay of one time step, accordingly to Equations~\ref{Arr:eq_start:standard_LSTM_forward}-\ref{Arr:eq_end:standard_LSTM_forward}. 
The cell state $\Bc$ also has a recurrent connection with one time step delay to himself via a multiplication with the forget gate  $\Bf$, and gets accumulated through a sum with the current input.
\label{Arr:fig:cellFB}}
\end{figure}

The central processing and storage unit for LSTM recurrent networks is
the {\em memory cell}.\index{LSTM!Memory Cell} As already mentioned, it avoids vanishing gradients and allows for
uniform credit assignment.
The most commonly used LSTM memory cell architecture in the 
literature \cite{Arr:Graves:05,Arr:Schmidhuber:15} 
contains forget gates \cite{Arr:Gers:99a,Arr:Gers:00}
and peephole connections \cite{Arr:Gers:00a}. 
In our previous work \cite{Arr:Hochreiter:01,Arr:Hochreiter:07}, 
we found that peephole connections are 
only useful for modeling time series, but not for 
language, meta-learning, or biological sequences. 
That peephole connections can be removed without performance decrease, 
was recently confirmed in a large assessment, where 
different LSTM architectures have been tested \cite{Arr:Greff:15}.
While LSTM networks are highly successful in various applications, 
the central memory cell architecture was not modified since 2000 \cite{Arr:Schmidhuber:15}.
A memory cell architecture without peepholes is depicted in
Figure~\ref{Arr:fig:cellFB}. 

In our definition of an LSTM network, all units of one kind are
pooled to a vector: $\Bz$ is the vector of cell input
activations, $\Bi$ is the vector of input gate
activations,  $\Bf$ is the vector of forget gate
activations,  $\Bc$ is the vector of memory cell states,
$\Bo$ is the vector of output gate
activations, and $\By$ is the vector of cell output 
activations.

We assume to have an input sequence, where the input vector at 
time $t$ is $\Bx_t$. The matrices $\BW_{\Bz}$, $\BW_{\Bi}$,
$\BW_{\Bf}$, and $\BW_{\Bo}$ correspond to the
weights of the connections between inputs and cell input, input gate,\index{LSTM!Gates} forget gate, and
output gate, respectively. The matrices $\BU_{\Bz}$, $\BU_{\Bi}$,
$\BU_{\Bf}$, and $\BU_{\Bo}$ correspond to the
weights of the connections between the cell output activations with one-step delay and cell input, input gate, forget gate, and
output gate, respectively. 
The vectors  $\Bb_{\Bz}$, $\Bb_{\Bi}$,
$\Bb_{\Bf}$, and $\Bb_{\Bo}$ are the bias vectors of cell input, input gate, forget gate, and
output gate, respectively.
The activation functions are $g$ for the cell input, $h$ for the cell
state, and $\sigma$ for the gates, where these functions are evaluated in a
component-wise manner if they are applied to vectors.
Typically, either the sigmoid $\frac{1}{1+\exp(-x)}$ or
$\tanh$ are used as activation functions.
$\odot$ denotes the point-wise multiplication
of two vectors. Without peepholes, the LSTM memory cell forward pass rules
are (see Figure~\ref{Arr:fig:cellFB}):
\begin{align}
\Bz_t \ &= \ g \left( \BW_{\Bz} \ \Bx_t \    + \ \BU_{\Bz} \ \By_{t-1} \  + \
   \Bb_{\Bz}\right) & \text{cell input} \label{Arr:eq_start:standard_LSTM_forward} \\
\Bi_t \ &= \ \sigma \left( \BW_{\Bi} \ \Bx_t \   + \ \BU_{\Bi} \ \By_{t-1} \  + \
    \Bb_{\Bi} \right) & \text{input gate} \\
\Bf_t \ &= \ \sigma \left( \BW_{\Bf} \ \Bx_t \  + \ \BU_{\Bf} \ \By_{t-1} \  + \
   \Bb_{\Bf} \right) & \text{forget gate} \\
\Bc_t \ &= \  \Bi_t \odot \Bz_t \ + \ 
\Bf_t \odot \Bc_{t-1} & \text{cell state} \\
\Bo_t \ &= \ \sigma \left( \BW_{\Bo} \ \Bx_t \   + \ \BU_{\Bo} \ \By_{t-1} \  + \
  \Bb_{\Bo} \right) & \text{output gate} \\
\By_t \ &= \ \Bo_t \odot h\left( \Bc_t \right) &
\text{cell output} \label{Arr:eq_end:standard_LSTM_forward}
\end{align}

\paragraph{Long-Term Dependencies vs.\ Uniform Credit Assignment.}
% ----------------------------------------------------------------------------------------------------------------------------------------------
The LSTM network has been proposed with the aim
to learn {\em long-term dependencies} in sequences
which span over long intervals
\cite{Arr:Hochreiter:97,Arr:Hochreiter:97e,Arr:Hochreiter:97f,Arr:Hochreiter:98}. 
However, besides extracting long-term dependencies, 
LSTM memory cells have another, even
more important, advantage in sequence learning:
as already described in the early 1990s,
LSTM memory cells allow for {\em uniform credit assignment}, that is,
the propagation of errors back to inputs without 
scaling them \cite{Arr:Hochreiter:91}. 
For uniform credit assignment of current LSTM architectures,
the forget gate $\Bf$ must be one or close to one.  
A memory cell without an input gate $\Bi$ just sums up all the squashed inputs it
receives during scanning the input sequence.
Thus, such a memory cell is equivalent to a unit that sees all sequence
elements at the same time, as has been shown via 
the ``Ersatzschaltbild'' (engl. equivalent circuit diagram) \cite{Arr:Hochreiter:91}.
If an output error occurs only at the end of the sequence,
such a memory cell, via backpropagation, supplies
the same delta error at the cell input unit $\Bz$ at every time
step.
Thus, all inputs obtain the same credit for producing the correct
output and are treated on an equal level and, consequently, the incoming weights to a memory cell 
are adjusted by using the same delta error at the input unit $\Bz$.

In contrast to LSTM memory cells, standard recurrent networks scale
the delta error and assign different credit to different inputs.
The more recent the input, the more credit it obtains.
The first inputs of the sequence are hidden from the final states of
the recurrent network.
In many learning tasks, however, important information is distributed over
the entire length of the sequence and can even occur at the very beginning. For
example, in language- and text-related tasks, 
the first words are often important for the meaning of a sentence. 
If the credit assignment is not uniform along the input sequence, then
learning is very limited. Learning would start by trying to improve
the prediction solely by using the most recent inputs.
Therefore, the solutions that can be found are restricted to those
that can be constructed if the last inputs are considered first.
Thus, only those solutions are found that are accessible by gradient
descent from regions in the parameter space that only use the most recent input information.
In general, these limitations lead to suboptimal solutions, since 
learning gets trapped in local optima. 
Typically, these local optima correspond to solutions 
which efficiently exploit the most recent information in the input
sequence, while information way back in the past is neglected.

\subsection{Layer-Wise Relevance Propagation (LRP)}
\label{Arr:sec:LRP}

Layer-wise relevance propagation (LRP) \cite{Arr:Bach:15} (cf.\ \cite{Arr:montavon2019overview}\index{Layer-Wise Relevance Propagation}\index{Explanation Methods!Layer-Wise Relevance Propagation}\index{Propagation-Based Explanations!Layer-Wise Relevance Propagation} for an overview) is a technique to explain individual predictions of deep neural networks in terms of input variables. For a given input and the neural network's prediction, it assigns a score to each of the input variables indicating to which extent they contributed to the prediction. LRP works by reverse-propagating the prediction through the network by means of heuristic propagation rules that apply to each layer of a neural network \cite{Arr:Bach:15}. In terms of computational cost the LRP method is very efficient, as it can be computed in one forward and backward pass through the network. In various applications LRP was shown to produce faithful explanations, even for highly complex and nonlinear networks used in computer vision \cite{Arr:Bach:15,Arr:Samek:TNNLS2017}. Besides it was able to detect biases in models and datasets used for training \cite{Arr:lapuschkin2019intelligent}, e.g.\ the presence of a copyright tag that spuriously correlated to the class `horse' in the Pascal VOC 2012 dataset. Further, it was used to get new insights in scientific and medical applications \cite{Arr:sturm2016eeg,Arr:Horst:SREP19,Arr:Yang:ICHI2018}, to interpret clustering  \cite{Arr:kauffmann2019}, to analyze audio data \cite{Arr:Thuillier:2018ICASSP,Arr:Becker:2018}, and to compare text classifiers for topic categorization \cite{Arr:Arras:PLOSONE2017}.

\paragraph{Conservative Propagation.}
\index{Layer-Wise Relevance Propagation!Conservation Principle}

LRP explains by redistributing the neural network output progressively from layer to layer until the input layer is reached. Similar to other works such as \cite{Arr:Landecker:13,Arr:Zhang:16,Arr:Shrikumar:PMLR2017}, the propagation procedure implemented by LRP is based on a local conservation principle: the net quantity, or relevance, received by any higher layer neuron is redistributed in the same amount to neurons of the layer below. 
In this way the relevance's flow is analog to the Kirchhoff’s first law for the conservation of electric charge, or to the continuity equation in physics for transportation in general form.
Concretely, if $j$ and $k$ are indices for neurons in two consecutive layers, and denoting by $R_{j \leftarrow k}$ the relevance flowing between two neurons, we have the equations:
\begin{align*}
&\textstyle \sum_j R_{j \leftarrow k} = R_k\\
R_j = &\textstyle \sum_k R_{j \leftarrow k}.
\end{align*}
This local enforcement of conservation induces conservation at coarser scales, in particular, conservation between consecutive layers $
\textstyle \sum_j R_j = \sum_j \sum_k R_{j \leftarrow k} = \sum_k \sum_j R_{j \leftarrow k} = \sum_k R_k$, and ultimately, conservation at the level of the whole deep neural network, i.e.\ given an input $\Bx = (x_i)_i$ and its prediction $f(\Bx)$, we have $\sum_i R_i \ = \ f(\Bx)$\footnote{The global conservation is exact up to the relevance absorbed by some stabilizing term, and by the biases, see details later in Section~\ref{Arr:sec:LRP_linear_mappings}.}. This global conservation property allows to interpret the result as the share by which each input variable has contributed to the prediction.

\paragraph{LRP in Deep Neural Networks.}

LRP has been most commonly applied to deep rectifier networks. In these networks, the activations at the current layer can be computed from activations in the previous layer as:
$$
\textstyle a_k = \max\big(0,\sum_{0,j} a_j w_{jk}\big)
$$
A general family of propagation rules for such types of layer is given by \cite{Arr:montavon2019overview}:
$$
R_j = \sum_k \frac{a_j \cdot \rho(w_{jk})}{\epsilon + \sum_{0,j} a_j \cdot \rho(w_{jk})} R_k
$$
\index{Layer-Wise Relevance Propagation!Propagation Rules}
Specific propagation rules such as LRP-$\epsilon$, LRP-$\alpha_1\beta_0$  and LRP-$\gamma$ fall under this umbrella. They are easy to implement \cite{Arr:Lapuschkin2016LRPtoolbox,Arr:montavon2019overview} and can be interpreted as the result of a deep Taylor decomposition of the neural network function \cite{Arr:Montavon:PR2017}.

On convolutional neural networks for computer vision, composite strategies making use of different rules at different layers have shown to work well in practice \cite{Arr:lapuschkin2017faces,Arr:montavon2019overview}. 
An alternative default strategy in computer vision is to uniformly employ the LRP-$\alpha_1\beta_0$ in every hidden layer \cite{Arr:Montavon:DSP18}, the latter has the advantage of having no free parameter, and delivers positive explanations.
On convolutional neural networks for text, LRP-$\epsilon$ with a small $\epsilon$ value  was found to work well \cite{Arr:Arras:PLOSONE2017,Arr:Poerner:ACL2018}, it provides a signed explanation.

\medskip

While LRP was described in the context of a layered feed-forward neural network, the principle is general enough to apply to arbitrary directed acyclic graphs, including recurrent neural networks unfolded in time such as LSTMs.

\section{Extending LRP for LSTMs}
\index{Layer-Wise Relevance Propagation!Extension to LSTMs}
We address the question of how to explain the LSTM model's output by expanding the previously described LRP technique to ``standard'' LSTM architectures, in the form they are most commonly used in the literature \cite{Arr:Greff:15}, i.e. following the recurrence Equations~\ref{Arr:eq_start:standard_LSTM_forward}-\ref{Arr:eq_end:standard_LSTM_forward} and Figure~\ref{Arr:fig:cellFB} introduced in Section~\ref{Arr:sec:LSTM}, and usually containing the {\it tanh} nonlinearity as an activation function for the cell input and the cell state.

For this, we first need to identify an appropriate structure of computation in these models, and introduce some notation. Let $s,g$ be the neurons representing the signal and the gate, let $p$ be the neuron representing the product of these two quantities. Let $f$ be the neuron corresponding to the forget gate. Let $k$ be the neuron on which the signal is being accumulated. Let $k-1, p-1,\dots$ be the same neurons at previous time steps. We can recompose the LSTM forward pass in terms of the following three elementary types of computation:
\begin{align*}
& \text{1.~linear mappings} & z_s &= \textstyle  \sum_{0,j} a_j w_{js} ~,~ z_g = \textstyle  \sum_{0,j} a_j w_{jg}\\
& \text{2.~gated interactions} & a_p &=  \tanh\big(z_s \big) \cdot \mathrm{sigm} \big( z_g \big)\\
&\text{3.~accumulation} & a_k &=  a_f \cdot a_{k-1} + a_p
\end{align*}

\begin{figure}
    \centering
    \includegraphics[width=0.7\textwidth]{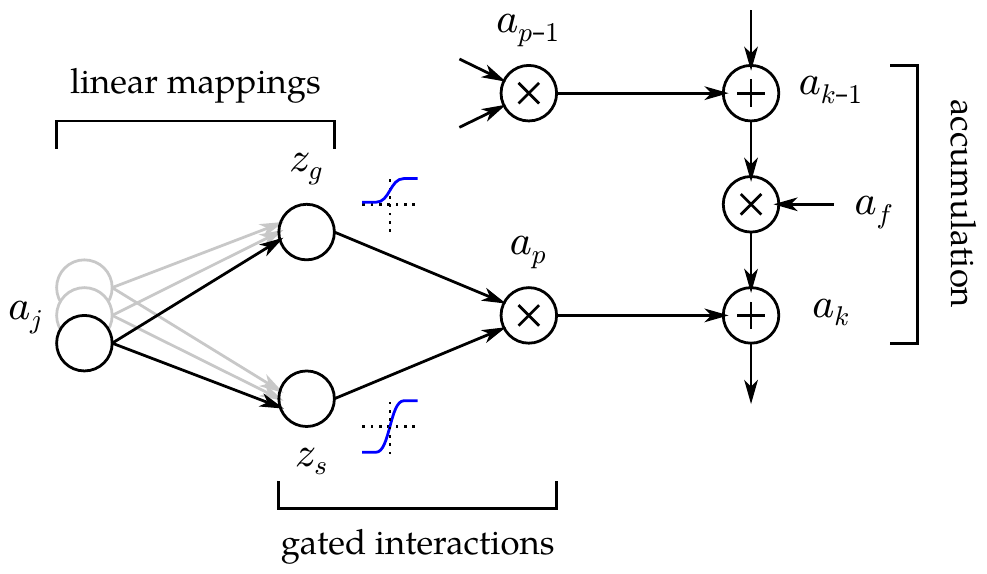}
    \caption{Three elementary computations performed by the LSTM from the perspective of LRP.}
    \label{Arr:fig:lrp}
\end{figure}

These three types of computation and the way they are typically interconnected are shown graphically in Figure~\ref{Arr:fig:lrp}.

Linear mappings form the input of the gated interactions. The output of some of the gated interactions enter into the accumulation function.

\subsection{Linear Mappings}
\label{Arr:sec:LRP_linear_mappings}

Each output of this computation is a weighted sum over a large number of input variables. Here, one strategy is to redistribute the relevance in {\it proportion} to the weighted activations $a_j w_{js}$, as they occur in the linear projection formulas above. One way of implementing this strategy is the epsilon-rule (LRP-$\epsilon$) given by \cite{Arr:Bach:15}:
$$
R_j = \sum_s \frac{a_j w_{js}}{\epsilon_s + \sum_{0,j} a_j w_{js}} R_s
$$
where $\epsilon_s  = \epsilon \cdot  \mathrm{sign}\big(\sum_{0,j} a_j w_{js}\big)$ is a small stabilizer that pushes the denominator away from zero by some constant factor, and  has the effect of absorbing some relevance when the weighted activations are weak or contradictory. This type of propagation rule was employed by previous works with recurrent neural networks \cite{Arr:Arjona-Medina:18,Arr:Arras:17,Arr:Ding:ACL2017,Arr:Poerner:ACL2018,Arr:Yang:ICHI2018}. A large value for $\epsilon$ tends to keep only the most salient factors of explanation.
Note that, in our notation, neuron biases are taken into account via a constant neuron $a_0 = 1$ whose connection weight is the corresponding bias. This neuron also gets assigned a share of relevance. However its relevance will not be propagated further and will get trapped in that neuron, since the ``bias neuron'' has no lower-layer connections.

\subsection{Gated Interactions}
\label{Arr:sec:LRP_gated_interactions}

These layers do not have a simple summing structure as the linear mappings. Their multiplicative nonlinearity makes them intrinsically more difficult to handle. Recently, three works extended the LRP propagation technique to recurrent neural networks, such as LSTMs \cite{Arr:Hochreiter:97} and GRUs \cite{Arr:Cho:EMNLP2014}, by proposing a rule to propagate the relevance through such product layers \cite{Arr:Arjona-Medina:18,Arr:Arras:17,Arr:Ding:ACL2017}. These LRP extensions were tested in the context of sentiment analysis, machine translation and reinforcement learning respectively. Arras et al. \cite{Arr:Arras:17}, in particular, proposed the signal-take-all redistribution rule
$$
(R_g,R_s) = (0,R_p)
$$
referred as ``LRP-all'' in our experiments. This redistribution strategy can be motivated in a similar way the gates were initially introduced in the LSTM model \cite{Arr:Hochreiter:97}: the gate units are intended to  control the flow of information in the LSTM, but not to be information themselves.

\medskip

\begin{figure}
    \centering
    \includegraphics[width=0.95\textwidth]{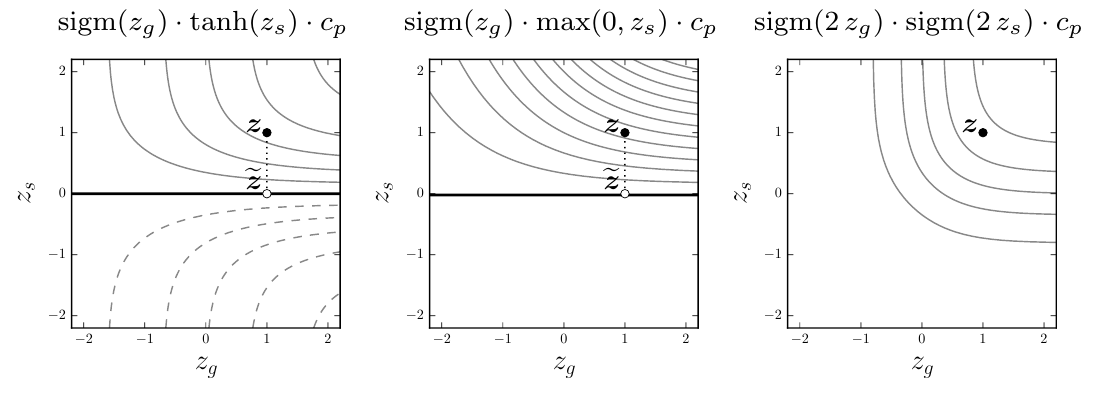}
    \vskip -3mm
    \caption{DTD relevance models for different choices of nonlinear functions with nearest root point (white dot). The left model is the standard LSTM. Positive contours are drawn as continuous lines, negative contours as dashed lines, and the dark line represents the zero-valued contour.}
    \label{Arr:fig:dtd}
\end{figure}

In the following, we provide further justification of this rule based on Deep Taylor Decomposition (DTD) \cite{Arr:Montavon:PR2017},\index{Layer-Wise Relevance Propagation!Deep Taylor Decomposition}\index{Theoretical Frameworks!Taylor Expansions!Deep Taylor Decomposition}\index{Deep Taylor Decomposition} a mathematical framework for analyzing the relevance propagation process in a deep network. DTD expresses the relevance obtained at a given layer as a function of the activations in the lower-layer, and determines how the relevance should be redistributed based on a Taylor expansion of this function.
Consider the relevance function $R_p(z_g,z_s)$ mapping the input $z=(z_g,z_s)$ of the gated interaction to the relevance received by the output of that module. We then write its Taylor expansion:
$$
R_p(z_g,z_s) = R_p(\widetilde{z}_g,\widetilde{z}_s)
+ \frac{\partial R_p}{\partial z_g}\Big|_{\widetilde{z}}
\cdot (z_g - \widetilde{z}_g)
+ \frac{\partial R_p}{\partial z_s}\Big|_{\widetilde{z}}
\cdot (z_s - \widetilde{z}_s)
+ \dots
$$
%o(\|z-\widetilde{z}\|)
where $\widetilde{z} = (\widetilde{z}_g,\widetilde{z}_s)$ is a root point of the function, and where the first-order terms can be used to determine on which lower-layer neurons ($g$ or $s$ the relevance should be propagated). In practice, a root point and its gradient are difficult to compute analytically. However, we can consider instead a relevance model \cite{Arr:Montavon:PR2017} which is easier to analyze, in our case, of the form:
$$
\widehat{R}_p(z_g,z_s) = \mathrm{sigm}(z_g) \cdot \tanh(z_s) \cdot c_p.
$$
The variable $c_p$ is constant and set such that ${R}_p(z_g,z_s)=\widehat{R}_p(z_g,z_s)$ locally. This model is a reasonable approximation when $R_p$ results from a propagation rule where the activation term naturally factors out (cf.\ \cite{Arr:Montavon:PR2017}). The relevance model for the gated interaction of the standard LSTM is depicted in Figure~\ref{Arr:fig:dtd} (left).

Having built the relevance model, we would like to perform a Taylor expansion of it at some root point in the vicinity of the observed point $z = (z_g,z_s)$. The nearest root point of the relevance model is found at $(\widetilde{z}_g,\widetilde{z}_s) = (z_g,0)$, and more generally any root point satisfies $\widetilde{z}_s=0$. A Taylor expansion of this simplified relevance model gives:
\begin{align*}
\widehat{R}_p(z_g,z_s) &= \widehat{R}_p(\widetilde{z}_g,\widetilde{z}_s)\\
& \hspace{5mm} + \mathrm{sigm}^\prime (\widetilde{z}_g) \cdot \tanh(\widetilde{z}_s) \cdot c_p \cdot (z_g - \widetilde{z}_g)  & ({} = R_g)\\
& \hspace{5mm} + \mathrm{sigm} (\widetilde{z}_g) \cdot \tanh'(\widetilde{z}_s) \cdot c_p \cdot (z_s - \widetilde{z}_s) &  ({} = R_s)\\
& \hspace{5mm} + \dots
\end{align*}
%o(\|z-\widetilde{z}\|).
Clearly, the first linear term $R_g$ is zero for the nearest root point, thus, no relevance will be redistributed to the gate, however, the saturation effect of the hyperbolic tangent can create a mismatch between the first-order term, and the function value to redistribute. However, if replacing in the LSTM the hyperbolic tangent by the identity or the ReLU nonlinearity (as this was done, for example, in \cite{Arr:rieger2018structuring}), then we get an
exact decomposition of the relevance model with $(R_g,R_s) = (0,\widehat{R}_p)$, since the Taylor remainder is exactly zero in this case.
This corresponds to the LRP-all redistribution rule.
\medskip

This section has justified the signal-take-all strategy for standard LSTMs. In Section \ref{Arr:sec:Explain_Adjusted_LSTM} modified LSTM variants that are tuned for further interpretability will benefit from a different propagation strategy. For example, using sigmoids both for the gate and the signal (cf. Figure~3 right) suggests a different propagation strategy. 

A more complete set of propagation rules that have been used in practice \cite{Arr:Arjona-Medina:18,Arr:Arras:17,Arr:Ding:ACL2017,Arr:Poerner:ACL2018,Arr:rieger2018structuring,Arr:Yang:ICHI2018}, and that we consider in our experiments, is given in Table~\ref{Arr:tab:product_rule_overview}.
In addition to the definitions provided in Table~\ref{Arr:tab:product_rule_overview}, in order to avoid near zero division, one may add a stabilizing term into the denominator of the LRP-prop and LRP-abs variants, similarly to the epsilon-rule stabilization for linear mappings. It has the form
$\epsilon \cdot  \mathrm{sign}\big(z_g + z_s\big)$ in the first case, and simply $\epsilon$  in the other case, where $\epsilon$ is a small positive number.

\begin{table}
\centering 
\caption{Overview of LRP propagation rules for gated interactions, and whether they derive from a deep Taylor decomposition. LRP-all stands for ``signal-take-all'', LRP-prop stands for ``proportional'', LRP-abs is similar to LRP-prop but with absolute values instead, and LRP-half corresponds to equal redistribution.}
\begin{tabular}{llllc}\toprule
 Name~~~~~~~~~&
 Proposed in~~~&
 Received by gate~~~~~&
 Received by signal~~~~~&
 DTD\\\midrule
LRP-all &
\cite{Arr:Arras:17} &
$R_g = 0$ &
$R_s = R_p$ &
$\checkmark$ \\
LRP-prop &
\cite{Arr:Ding:ACL2017,Arr:Arjona-Medina:18}&
$R_g = \tfrac {z_g } {z_g + z_s } R_p$ &
$R_s = \tfrac {z_s } {z_g + z_s } R_p$ &
$\times$ \\
LRP-abs &
{}&
$R_g = \tfrac {|z_g| } {|z_g| + |z_s| } R_p $ &
$R_s = \tfrac {|z_s| } {|z_g| + |z_s| } R_p $  &
$\times$ \\
LRP-half &
\cite{Arr:Arjona-Medina:18} &
$R_g = 0.5 \cdot R_p $ &
$R_s = 0.5 \cdot R_p $ &
$\times$ \\
\bottomrule
\end{tabular}
\label{Arr:tab:product_rule_overview}
\end{table}

\subsection{Accumulation}
\label{Arr:sec:LRP_accumulation}

The last type of module one needs to consider is the accumulation module that discounts the LSTM memory state with a ``forget'' factor, and adds a small additive term based on current observations:
$$
a_k = a_f \cdot a_{k-1} + a_p.
$$
Consider the relevance $R_k$ of the accumulator neuron $a_k$ for the final time step. Define $R_k = a_k \cdot c_k$. 
Through the accumulation module, we get the following redistribution:
\begin{align*}
\allowdisplaybreaks
R_{p} &= a_p \cdot c_{k} \\
R_{k-1} &= a_f \cdot a_{k-1} \cdot c_{k},
\end{align*}
where we have used the signal-take-all strategy in the product, and the epsilon-rule (with no stabilizer) in the sum.
Iterating this redistribution process on previous time steps, we obtain:
\begin{align*}
\allowdisplaybreaks
R_{p-1} &= a_f \cdot a_{p-1} \cdot c_k\\
R_{p-2} &= (a_f \cdot a_{f-1}) \cdot a_{p-2} \cdot c_k\\[-2mm]
&~\,\vdots\\[-2mm]
R_{p-T} &= \textstyle \big(\prod_{t=1}^T a_{f-t+1}\big) \cdot a_{p-T} \cdot c_k.
\end{align*}
Note that, here, we assume a simplified recurrence structure over the standard LSTM presented in Figure~\ref{Arr:fig:cellFB}, in particular we assume that neurons $a_p$ do not redistribute relevance to past time steps via $z_s$ (i.e. $z_s$ is connected only to the current input and not to previous recurrent states), to simplify the present analysis.

Now we inspect the structure of the relevance scores $R_p,\dots,R_{p-T}$ at each time step, as given above. We can see that the relevance terms can be divided into three parts:
\begin{enumerate}
\item A product of forget gates: This term tends to decrease exponentially with every further redistribution step, unless the forget gate is equal to one. In other words, only the few most recent time steps will receive relevance.
\item The value of the product neuron  $a_p$ at the current time step. In other words, the relevance at a given time step is directly influenced by the activation of its representative neuron, which can be either positive or negative.
\item A term that does not change with the time steps, and relates to the amount of relevance available for redistribution.
\end{enumerate}
These observations on the structure of the relevance over time provide a further justification for the LRP explanation procedure, which we will be validating empirically in Section~\ref{Arr:sec:experiments}. 
They also serve as a starting point to propose new variants of the LSTM for which the relevance redistribution satisfies further constraints, as proposed in the following Section~\ref{Arr:sec:Explain_Adjusted_LSTM}.

\section{LSTM Architectures Motivated by LRP}
\label{Arr:sec:Explain_Adjusted_LSTM}
% ----------------------------------------------------------------------------------------------------------------------------------------------
%% Short intro
\index{LSTM!Explainable LSTMs}
A ``standard'' LSTM network with fully connected LSTM blocks, as presented in Figure~\ref{Arr:fig:cellFB}, is a very powerful network capable of modelling extremely complex sequential tasks. However, in many cases, an LSTM network with a reduced complexity is able to solve the same problems with a similar prediction performance.
With the further goal of increasing the model's interpretability, we propose some modifications which simplify the LSTM network and make the resulting model easier to explain with LRP.
%%%%
\subsection{LSTM for LRP Backward Analysis: Nondecreasing Memory Cells}

% Short description
The LRP backward propagation procedure is made simpler if memory cell states $\Bc_t$ are nondecreasing,
this way the contribution of each input to each memory cell
is well-defined, and the problem that a 
negative and a positive contribution cancel each other is avoided. 
For nondecreasing memory cells 
and backward analysis with LRP, 
we make the following assumptions over the LSTM network from Figure~\ref{Arr:fig:cellFB} and Equations~\ref{Arr:eq_start:standard_LSTM_forward}-\ref{Arr:eq_end:standard_LSTM_forward}:
\begin{enumerate}[label=\textbf{(A\arabic*)}]
\item $\Bf_t=1$ for all $t$. That is, the forget gate is always 1 and
  nothing is forgotten. This ensures uniform credit assignment over time,
  and alleviates the problem identified earlier in Section~\ref{Arr:sec:LRP_accumulation}
  that the relevance redistributed via the accumulation module decreases over time.

\item $g>0$, that is, the cell input activation function $g$ is positive. For example we can use a sigmoid
  $\sigma(x)= \frac{1}{1+\exp(-x)}$: $g(x)=a_g \sigma(x)$, with
  $a_g \in \{2,3,4\}$.
  Indeed methods like LRP and the epsilon-rule for linear mappings (cf. Section~\ref{Arr:sec:LRP_linear_mappings}) face numerical stability issues when negative contributions
  cancel with positive contributions \cite{Arr:Montavon:DSP18}.
  With a positive $g$ all
  contributions are positive, and the redistribution in the LSTM accumulation module is made more stable.
  Further, we assume that the cell input $\Bz$ has a negative bias, that is,
  $\Bb_{\Bz}<0$. This is important to avoid the drift effect.
  The drift effect is that the memory content only gets positive
  contributions, which leads to an increase of $\Bc$ over time.
  Typical values are $\Bb_{\Bz} \in \{-1,-2,-3,-4,-5\}$.

\item We want to ensure that for the cell state activation it holds $h(0)=0$, such that, if the memory content is zero,
  then nothing is transferred to the next layer.
  Therefore we set $h=a_h \tanh$, with $a_h \in \{1,2,4\}$.

\item The cell input $\Bz$ is only connected to the input, and is not connected
  to other LSTM memory cell outputs.  Which means $\BU_{\Bz}$ is zero. This ensures
  that LRP assigns relevance $\Bz$ to the input and $\Bz$ is not
  disturbed by redistributing relevance to the network.

\item The input gate $\Bi$ has only connections to other memory cell outputs, and is not connected to the input. That is,
  $\BW_{\Bi}$ is zero. This ensures
  that LRP assigns relevance only via $\Bz$ to the input.

\item The output gate $\Bo$ has only connections to other memory cell outputs, and is not connected to the input. That is,
  $\BW_{\Bo}$ is zero. This ensures
  that LRP assigns relevance only via $\Bz$ to the input.

\item The input gate $\Bi$ has a negative bias, that is,
  $\Bb_{\Bi}<0$. Like with the cell input the negative bias
  avoids the drift effect.
  Typical values are $\Bb_{\Bi} \in \{-1,-2,-3,-4\}$.

\item The output gate $\Bo$ may also have a negative bias, that is,
  $\Bb_{\Bo}<0$. This allows to bring in different memory cells at
  different time points. It is related to resource allocation.
  
\item The memory cell state content is initialized with zero at time $t=0$,
  that is, $\Bc_0=0$. Lastly, the memory cell content $\Bc_t$ 
  is non-negative $\Bc_t \geq 0$ for all $t$, since 
  $\Bz_t \geq 0$ and $\Bi_t \geq 0$.
 
\end{enumerate}

\begin{figure}[htb]
\centering
\includegraphics[angle=0,width=1.0\textwidth]{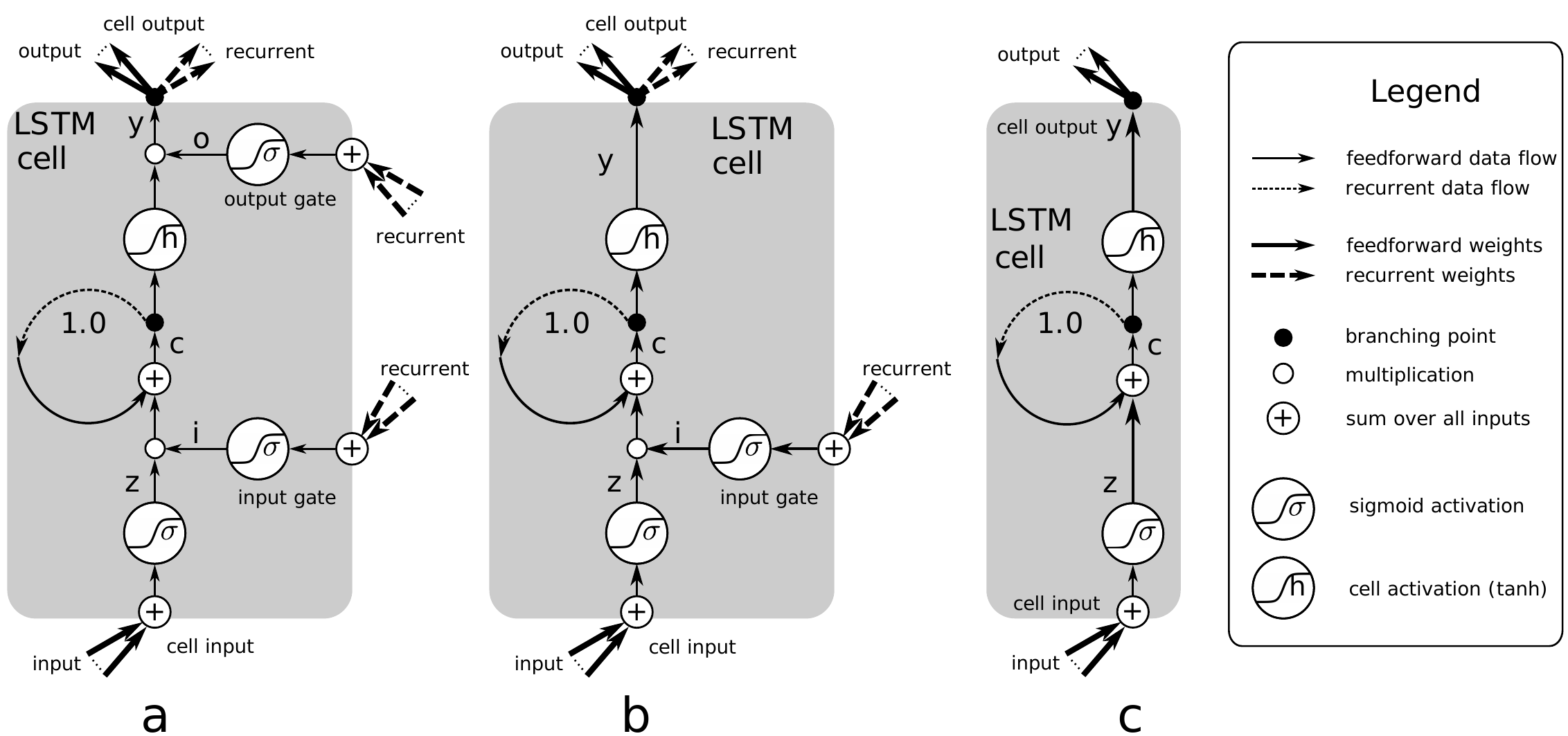}
\caption{LSTM memory cell used for Layer-Wise Relevance Propagation (LRP).\newline 
(a) $\Bz$ is the vector of cell input
activations, $\Bi$ is the vector of input gate
activations,  $\Bc$ is the vector of memory cell states,
$\Bo$ is the vector of output gate
activations, and $\By$ is the vector of cell output 
activations. 
(b) The memory cell is nondecreasing and guarantees the Markov
property.
(a and b) Data flow is either ``feed-forward''
without delay or ``recurrent'' with a one-step delay.
External input reaches the LSTM network 
only via the cell input $\Bz$. All gates only receive
recurrent input, that is, from other memory cell outputs.
(c) LSTM memory cell without gates. External input is stored in the memory cell via the
input $\Bz$.
\label{Arr:fig:3lstm}}
\end{figure}

The resulting LSTM forward pass rules for LRP are:
\begin{align}
\Bz_t \ &= \ a_g \ \sigma \left( \BW_{\Bz} \ \Bx_t \ + \
   \Bb_{\Bz}\right) & \text{cell input} \\
\Bi_t \ &= \ \sigma \left( \BU_{\Bi} \ \By_{t-1}\ + \
    \Bb_{\Bi} \right) & \text{input gate} \\
\Bc_t \ &= \  \Bi_t \ \odot \ \Bz_t \ + \ \Bc_{t-1} & \text{cell state} \\
\Bo_t \ &= \ \sigma \left( \BU_{\Bo} \ \By_{t-1} \ + \
  \Bb_{\Bo} \right) & \text{output gate} \\
\By_t \ &= \ \Bo_t \ \odot \ a_h \ \tanh\left( \Bc_t \right) &
\text{cell output}
\end{align}
See Figure~\ref{Arr:fig:3lstm}a which depicts these  
forward pass rules for LRP. 

\subsection{LSTM for LRP Backward Analysis: Keeping the Markov Property}

Forget gates can modify the memory cells' information at some future time step, i.e. they could completely erase the cells' state content. 
Output gates can hide the cells' information and deliver it in the future, i.e. output gates can be closed and open at some future time, masking all information stored by the cell.
Thus, in order to guarantee the Markov property, the forget and output gates must be disconnected.

The resulting LSTM forward pass rules for Markov memory cells are:
\begin{align}
\Bz_t \ &= \ a_g \ \sigma \left( \BW_{\Bz} \ \Bx_t \ + \
   \Bb_{\Bz}\right) & \text{cell input} \\
\Bi_t \ &= \ \sigma \left( \BU_{\Bi} \ \By_{t-1} \ + \
    \Bb_{\Bi} \right) & \text{input gate} \\
\Bc_t \ &= \  \Bi_t \ \odot \ \Bz_t \ + \ \Bc_{t-1} & \text{cell state} \\
\By_t \ &= \ a_h \ \tanh\left( \Bc_t \right) &
\text{cell output}
\end{align}
See Figure~\ref{Arr:fig:3lstm}b for an LSTM memory cell that guarantees the Markov property.

\subsection{LSTM without Gates}

The most simple LSTM architecture for backward analysis does not
use any gates. Therefore complex dynamics that have to be treated 
in the LRP backward analysis are avoided.

The resulting LSTM forward pass rules are:
\begin{align}
\Bz_t \ &= \ a_g \ \sigma \left( \BW_{\Bz} \ \Bx_t \ + \
   \Bb_{\Bz}\right) & \text{cell input} \\
\Bc_t \ &= \  \Bz_t \ + \ \Bc_{t-1} & \text{cell state} \\
\By_t \ &= \ a_h \ \tanh\left( \Bc_t \right) &
\text{cell output}
\end{align}
Note that even this simple architecture can solve sequential problems, since different biases can be learned by the cell inputs to specialize on different time steps and activate the memory cell output accordingly.

See Figure~\ref{Arr:fig:3lstm}c for an LSTM memory cell without
gates which perfectly redistributes the relevance across the input sequence.

%\FloatBarrier

\section{Experiments}
\label{Arr:sec:experiments}
% ----------------------------------------------------------------------------------------------------------------------------------------------

\subsection{Validating Explanations on Standard LSTMs: Selectivity and Fidelity}
\label{Arr:sec:Experiments_with_standard_LSTMs}
% ----------------------------------------------------------------------------------------------------------------------------------------------
\index{Evaluating Explanations}
\index{Evaluating Explanations!Perturbation Analysis}
First we verify that the LRP explanation is able to select input positions that are the most determinant either in {\it supporting} or in {\it contradicting} an LSTM's prediction, using a sentiment prediction task. 
To that end we perform a perturbation experiment aka ``pixel flipping'' or ``region perturbation'' \cite{Arr:Bach:15,Arr:Samek:TNNLS2017} commonly used in computer vision to evaluate and generate explanations, e.g. \cite{Arr:Lundberg:NIPS2017,Arr:Ancona:ICLR2018,Arr:Chen:ICML2018,Arr:Morcos:ICLR2018}.
Here we confirm whether the {\it sign} and {\it ordering} of the relevance reflect what the LSTM considers as highly speaking {\it for} or {\it against} a particular class.

\index{Evaluating Explanations!Toy Task}
Another property of the relevance we test is fidelity. To that end we use a synthetic task where the input-output relationship is known and compare the relevances w.r.t some ground truth explanation.
By using a synthetic toy task we can avoid problems of disentangling errors made by the model from errors made by the explanation \cite{Arr:Sundararajan:ICML2017}.
Here we seek to validate the {\it magnitude} of the relevance as a continuously distributed quantity. 
To the best of our knowledge we are the first one to conduct such a continuous analysis of the relevance in recurrent neural networks.
Yet another work validated LSTM explanations via a toy classification task \cite{Arr:Yang:ICHI2018}, however it practically treated the relevance as a binary variable.

\paragraph{Explanation Methods.}
% ---------------------------------------------------------------------------------------------------------------------------------------------
Now we introduce the various explanation methods we consider in our experiments with standard LSTMs.
For the LRP explanation technique we consider all product rule variants specified in Table~\ref{Arr:tab:product_rule_overview} (cf. Section~\ref{Arr:sec:LRP_gated_interactions}) , i.e. LRP-all \cite{Arr:Arras:17}, LRP-prop \cite{Arr:Ding:ACL2017,Arr:Arjona-Medina:18}, LRP-abs and LRP-half \cite{Arr:Arjona-Medina:18}.
Since the LRP backward pass delivers one relevance value per input dimension, we simply sum up the relevances across the input dimensions to get one relevance value per time step. 
Besides LRP we also consider gradient-based explanation \cite{Arr:Gevrey:2003,Arr:Li:NAACL2016,Arr:Simonyan:ICLR2014,Arr:Denil:2015}, occlusion-based relevance \cite{Arr:Li:ArXiv2017,Arr:Zeiler:ECCV2014}, and Contextual Decomposition (CD) \cite{Arr:Murdoch:ICLR2018}, as alternative methods.

For the gradient-based explanation we use as the relevance, either the prediction function's partial derivative w.r.t.\ the input dimension of interest and square this quantity, we denote this variant simply as {\it Gradient}, or else we multiply this derivative by the input dimension's value, we call it {\it Gradient $\times$ Input} relevance. In both cases, similarly to LRP, the relevance of several input dimensions can be summed up to obtain one relevance value per time step.

For the occlusion-based explanation we take as the relevance, either a difference of prediction function values (i.e. of prediction scores before softmax normalization), we denote this variant as {\it Occlusion}$_\text{f-diff}$, or else we use a difference of predicted probabilities, we call it {\it Occlusion}$_\text{P-diff}$, where the difference is calculated over the model's prediction on the original input and a prediction with an altered input where the position of interest (for which the relevance is being computed) is set to zero.

For the CD explanation method \cite{Arr:Murdoch:ICLR2018} we employ the code from the authors\footnote{\url{https://github.com/jamie-murdoch/ContextualDecomposition}} to generate one relevance value per time step (this necessitates to run the CD decomposition as many times as there are time steps in the input sequence).

\paragraph{Testing Selectivity.}
% ---------------------------------------------------------------------------------------------------------------------------------------------
\index{Explanations!Desiderata of Explanations!Selectivity}
In order to assess the selectivity, we consider a five-class sentiment prediction task of movie reviews.
As a dataset we use the Stanford Sentiment Treebank (SST) \cite{Arr:Socher:EMNLP2013} which contains labels (from very negative, negative, neutral, positive, to very positive sentiment) for resp.\ 8544/1101/2210 train/val/test sentences and their constituent phrases.
As an LSTM model we employ the bidirectional LSTM model from Li et al. \cite{Arr:Li:NAACL2016} already trained on SST\footnote{\url{https://github.com/jiweil/Visualizing-and-Understanding-Neural-Models-in-NLP}} and previously employed by the authors to perform a gradient-based analysis on the network decisions.
The input consists of a sequence of 60-dimensional word embeddings, the LSTM hidden layer has size 60, and the only text preprocessing is lowercasing. 
On binary sentiment classification of full sentences (ignoring the neutral class) the model reaches 82.9\% test accuracy, and on five-class sentiment prediction of full sentences it achieves 46.3\% accuracy.

For the perturbation experiment we consider all test sentences with a length of at least ten words (thus we retain 1849 sentences), 
and compute word-level relevances (i.e. one relevance value per time step) using as the target output class the {\it true} sentence class, and considering all five classes of the sentiment prediction task.
For the computation of the LRP relevance we use as a stabilizer value, for linear mappings and product layers, $\epsilon=0.001$\footnote{Except for the LRP-prop variant, where we take $\epsilon=0.2$. We tried following values: [0.001, 0.01, 0.1, 0.2, 0.3, 0.4, 1.0], and took the lowest one to achieve numerical stability.}.

Then, given these word-level relevances, we iteratively remove up to five words from each input sentence, either in {\it decreasing} or {\it increasing} order of their relevance, depending on whether the corresponding sentence was initially correctly or falsely classified by the LSTM. We expect this input modification to decrease resp. increase the model's confidence for the true class, which we measure in terms of the model's accuracy. 
For removing a word we simply discard it from the input sequence and concatenate the remaining parts of the sentence. An alternative removal scheme would have been to set the corresponding word embedding to zero in the input (which in practice gave us similar results), however the former enables us to generate more natural texts, although we acknowledge that the resulting sentence might be partly syntactically broken as pointed out by Poerner et al. \cite{Arr:Poerner:ACL2018}.

\begin{figure}[th!]
	\centering
	\includegraphics[width=0.99\columnwidth]{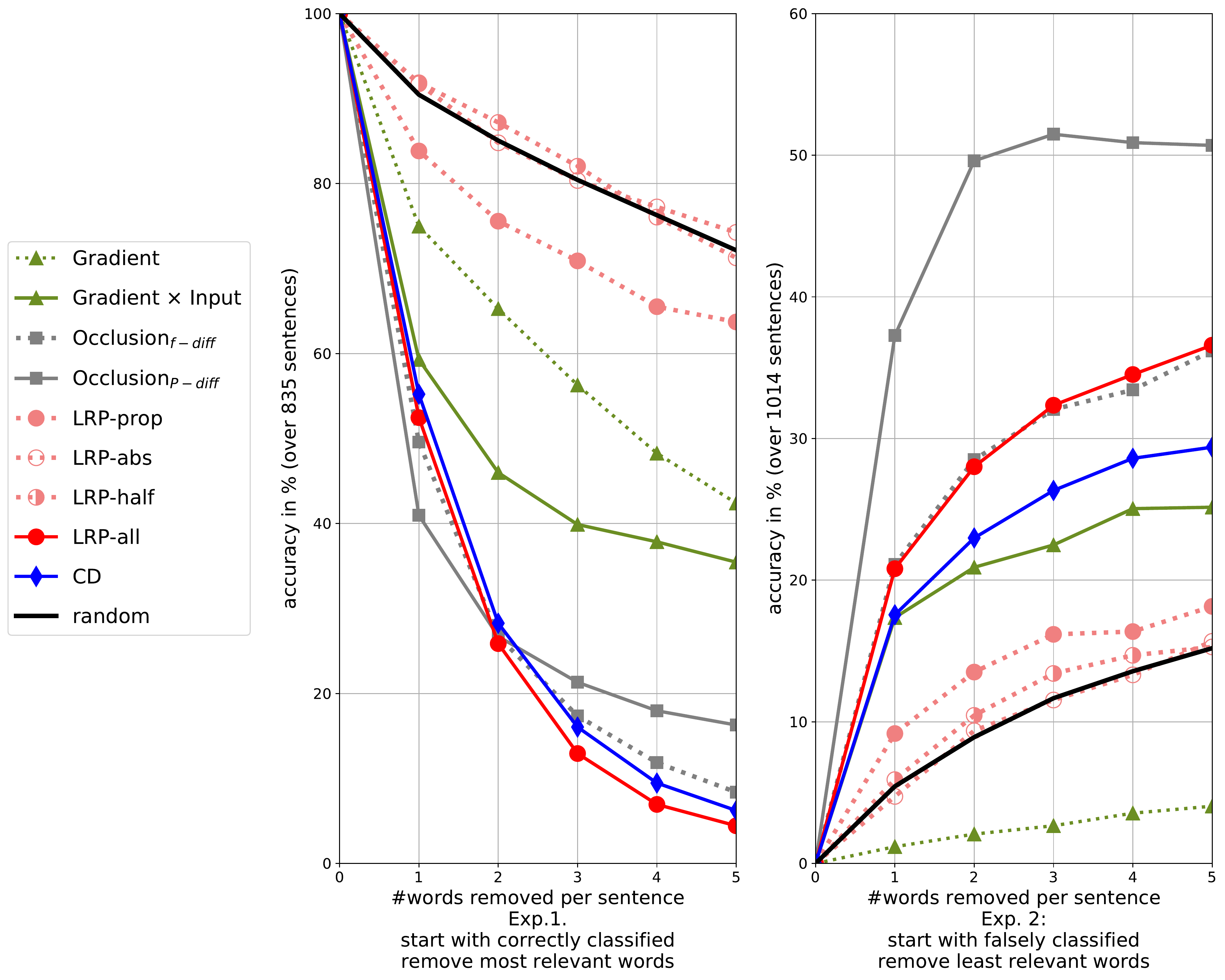}
	\caption{Impact of word removal on initially correctly (left) and initially falsely (right) classified sentences. The relevance target class is the true sentence class, and words are deleted in decreasing (left) and increasing (right) order of their relevance. Random deletion is averaged over 10 runs (std $<$ 0.02). A steep decline (left) and incline (right) indicate selective relevance.}
	\label{Arr:fig:selectivity_experiment}
\end{figure}

Our results of the perturbation experiment are compiled in Figure~\ref{Arr:fig:selectivity_experiment}.

When looking at the removal of the most relevant words (Figure~\ref{Arr:fig:selectivity_experiment} left), we observe that the occlusion-based relevance, LRP-all and CD are the most competitive methods, and perform on-par;
followed by  {\it Gradient $\times$ Input}, which performs better than {\it Gradient}.
The remaining methods, which are the other LRP variants LRP-prop, LRP-abs, LRP-half are almost equivalent to random, and thus not adequate to detect words speaking {\it for} a specific class.

In the removal of the least relevant words (Figure~\ref{Arr:fig:selectivity_experiment} right), {\it Occlusion}$_\text{P-diff}$ performs best; followed by {\it Occlusion}$_\text{f-diff}$, LRP-all, {CD} and {\it Gradient $\times$ Input}.
Again the remaining LRP variants are almost equivalent to random. However this time {\it Gradient} performs worse than random, this indicates that low {\it Gradient} relevance is more likely to identify unimportant words for the classification problem (like stop-words) rather than identifying words speaking {\it against} a specific class (this was also observed in previous work, see e.g. \cite{Arr:Arras:17} Table 1).

In summary, our perturbation experiment in sentiment analysis suggests that if one is interested in identifying the most influential positions that strongly {\it support} or {\it inhibit} a specific classification decision using a standard LSTM model,
then the occlusion-based relevance, the LRP method with the product rule from Arras et al. \cite{Arr:Arras:17}, and the CD method of Murdoch et al. \cite{Arr:Murdoch:ICLR2018} are good candidates to provide this information.

For another evaluation of recurrent neural networks explanations, including a standard LSTM model, we further refer to Poerner et al. \cite{Arr:Poerner:ACL2018}, in particular to their experiment using a subject-verb agreement task. 
Here the authors find that LRP and DeepLIFT \cite{Arr:Shrikumar:PMLR2017} perform best among the tested explanation methods, both when using the signal-take-all strategy proposed in Arras et al. \cite{Arr:Arras:17} 
for product layers\footnote{Ancona et al. \cite{Arr:Ancona:ICLR2018} also performed a comparative study of explanations on LSTMs, however, in order to redistribute the relevance through product layers, the authors use standard gradient backpropagation. This redistribution scheme violates one of the key underlying property of LRP, which is local relevance conservation, hence their results for LRP are not conclusive.}.

\paragraph{Testing Fidelity.}
% ----------------------------------------------------------------------------------------------------------------------------------------------
\index{Explanations!Desiderata of Explanations!Faithfulness}
In order to validate the fidelity, we consider a toy task with a linear input-output relationship. In particular we use the addition/subtraction of two numbers.
Accordingly we expect the relevances to be linearly related to the actual input values, which we can directly measure in terms of the empirical correlation.

For our purpose we use a variant of the adding problem of Hochreiter et al. \cite{Arr:Hochreiter:97e}, where instead of using explicit markers, we use implicit ones; further we remove the sequence start and end positions. 
This way we enforce the LSTM model to attribute non-zero relevance {\it only} to the relevant numbers in the input and not to markers (since in general it is unclear what ``ground truth'' relevance should be attributed to a marker, to which we could compare the computed relevance to).
Thus our input sequence of length $T$ has the form:
\begin{equation*}
\scriptsize
   \begin{bmatrix} 
   n_{1} 	& 0  	\\
   ...    	& 0 	\\
   n_{a-1}	& 0  	\\
   0     	& n_a  	\\
   n_{a+1} 	& 0 	\\
   ... 		& 0 	\\
   n_{b-1}	& 0 	\\
   0     	& n_b  	\\
   n_{b+1} 	& 0 	\\
   ... 		& 0 	\\
   n_{T}	& 0 	\\
   \end{bmatrix}  
\end{equation*}
where the non-zero entries  $n_t$, with $t \in \{1,...,T\}$, are randomly sampled real numbers, and the two relevant positions $a$ and $b$ are sampled uniformly among $\{1,...,T\}$ with $a<b$.
The target output is  $n_a+n_b$ for addition, and $n_a-n_b$ for subtraction.
To ensure that the train/val/test sets do not overlap, we use 10000 sequences with $T \in \{4,...,10\}$ for training, 2500 sequences with $T \in \{11,12\}$ for validation, and 2500 sequences with $T \in \{13,14\}$ as test set. Training is performed by minimizing Mean Squared Error (MSE).

More particularly, we consider two {\it minimal} tasks, which are solvable by a standard LSTM with only {\it one} memory cell, followed by a linear output layer with no  bias (thus the model has 17 trainable parameters):
\begin{itemize}
 \item the addition of {\it signed} numbers (where $n_t$ is sampled uniformly from $[-1,-0.5] \cup [0.5, 1.0]$),
 \item the subtraction of {\it positive} numbers (where $n_t$ is sampled uniformly from $[0.5, 1.0]$)\footnote{We use an arbitrary minimum magnitude of 0.5 only to simplify training (since sampling very small numbers would encourage the model weights to grow rapidly).}.
\end{itemize}
For each task we train 50 LSTM models with a validation MSE $<10^{-4}$, the resulting test MSE is also $<10^{-4}$.

Then, using the model's predicted output, we compute one relevance value $R_t$ per time step $t \in \{1,...,T\}$, for each considered explanation method.

For the occlusion-based relevance we use only the {\it Occlusion}$_\text{f-diff}$ variant, since the model output is one-dimensional and the considered task is a regression.
For the gradient-based relevance we report only the {\it Gradient $\times$ Input} results, since the pure {\it Gradient} performs very poorly.
For the computation of the LRP relevance we didn't find it necessary to add any stabilizing term (therefore we use $\epsilon=0.0$ for all LRP rules).

Our results are reported in Table~\ref{Arr:tab:faithfulness_experiment}. 
For the positions $a$ and $b$, we checked whether there is a correlation between the computed relevance and the input numbers' actual value. Besides, we verified the portion of the relevance (in absolute value) assigned to these  positions, compared to the relevance attributed to all time steps in the sequence.

\begin{table}[t!]
	\centering \footnotesize 
	\caption{Statistics of the relevance w.r.t. the numbers $n_a$ and $n_b$ on toy arithmetic tasks. $\rho$ denotes the correlation and $E$ the mean and for each LSTM model these statistics are computed over 2500 test points. Reported results are the mean (and standard deviation in parenthesis), in \%, over 50 trained LSTM models.}	
	\setlength{\tabcolsep}{6pt}
	\begin{tabular}{ m{2.6cm} m{2.2cm} m{2.2cm} m{2.2cm}}\\
		{}    			                & {$\rho{(n_a, R_a)}$} 	                    & {$\rho{(n_b, R_b)}$}                      & { $ E[\frac{\vert {R_a} \vert + \vert {R_b} \vert}{\sum_{t}{ \vert R_t} \vert } ]$ }      \\[3mm]\midrule
		
		           & \multicolumn{3}{l}{Addition $n_a+n_b$}                                                                                                                                            \\\midrule
		Gradient $\times$ Input 	    & \textbf{99.960} {\tiny(0.017)}			& \textbf{99.954} {\tiny(0.019)}			& \textbf{99.68} {\tiny(0.53)}				                                                \\	
		Occlusion 				        & \textbf{99.990} {\tiny(0.004)}			& \textbf{99.990} {\tiny(0.004)}			& \textbf{99.82} {\tiny(0.27)}		                                                        \\	 
		{LRP-prop}		                & 0.785 {\tiny(3.619)}			            & 10.111 {\tiny(12.362)}				    & 18.14 {\tiny(4.23)}				            	                                        \\	 
		{LRP-abs}		                & 7.002 {\tiny(6.224)}			            & 12.410 {\tiny(17.440)} 				    & 18.01 {\tiny(4.48)}				                                                        \\	 
		{LRP-half}		                & 29.035 {\tiny(9.478)}				        & 51.460 {\tiny(19.939)} 				    & 54.09 {\tiny(17.53)}								                                        \\	 
		{LRP-all}		                & \textbf{99.995} {\tiny(0.002)}			& \textbf{99.995} {\tiny(0.002)}		    & \textbf{99.95} {\tiny(0.05)}			                                                    \\	
		CD				            	& \textbf{99.997} {\tiny(0.002)}			& \textbf{99.997} {\tiny(0.002)}			& \textbf{99.92} {\tiny(0.06)}		                                                        \\ \\[-3mm] \midrule
	
		           & \multicolumn{3}{l}{Subtraction $n_a-n_b$ }                                                                                                                                        \\\midrule
		Gradient $\times$ Input 		& \textbf{97.9} {\tiny(1.6)}	            & \textbf{-98.8} {\tiny(0.6)}         	    & \textbf{98.3} {\tiny(0.6)}					                                            \\	
		Occlusion 				        & 99.0 {\tiny(2.0)}			                & -69.0 {\tiny(19.1)}					    & 25.4 {\tiny(16.8)} 						                                                \\	 
		{LRP-prop}		                & 3.1 {\tiny(4.8)} 					        & -8.4 {\tiny(18.9)}			            & 15.0 {\tiny(2.4)}						                                                    \\
		{LRP-abs}		                & 1.2 {\tiny(7.6)} 					        & -23.0 {\tiny(11.1)}					    & 15.1 {\tiny(1.6)}								                                            \\
		{LRP-half}	                    & 7.7 {\tiny(15.3)} 			            & -28.9 {\tiny(6.4)}			            & 42.3 {\tiny(8.3)}						                                                    \\
		{LRP-all}		                & \textbf{98.5} {\tiny(3.5)}			    & \textbf{-99.3} {\tiny(1.3)}	            & \textbf{99.3} {\tiny(0.6)}				                                                \\	 
		CD					            & -25.9 {\tiny(39.1)}					    & -50.0 {\tiny(29.2)}					    & 49.4 {\tiny(26.1)} 						                                                \\	 
	\end{tabular}

	\label{Arr:tab:faithfulness_experiment}
\end{table}

Interestingly several methods pass our sanity check (they are reported in bold in the Table) and attribute as expected a correlation of almost one in the addition task, namely: {\it Gradient $\times$ Input}, {\it Occlusion}, LRP-all and {CD}.

However, on subtraction, only {\it Gradient $\times$ Input} and LRP-all assign a correct correlation of near one to the first number, and of near minus one to the second number, while the remaining explanation methods fail completely.

For both addition and subtraction, we observe that methods that fail in the correlation results also erroneously assign a non-negligible portion of the relevance to clearly unimportant time steps.

One key difference between the two arithmetic tasks we considered, is that our addition task is non-sequential and solvable by a Bag-of-Words approach (i.e. by ignoring the ordering of the inputs), while our subtraction task is truly sequential and requires the LSTM model to remember which number arrives in the first position and which number in the second.

From this sanity check, we certainly can not deduce that every method that passes the subtraction test is also appropriate to explain {\it any} complex nonlinear prediction task.
However, we postulate that an explanation method that fails on such a test with the smallest possible number of free parameters (i.e. an LSTM with one memory cell) is generally a less suited method.

In this vein, our present analysis opens up new avenues for improving and testing LSTM explanation methods in general, including the LRP method and its LRP-all variant, whose results in our arithmetic task 
degrade when more memory cells are included to the LSTM model, which suggests that gates might also be used by the standard LSTM to store the input numbers' value\footnote{The same phenomenon can occur, on the addition problem, when using only positive numbers as input. Whereas in the specific toy tasks we considered, the cell input ($z_t$) is required to process the numbers to add/subtract, and the cell state ($c_t$) accumulates the result of the arithmetic operation.}.
The latter phenomenon could be either avoided by adapting the LSTM architecture, or could be taken into account in the relevance propagation procedure by employing alternative propagation rules for products. 
This leads us to the next subsection, where we use an adapted LSTM model and different product rules, for the task of reward redistribution.

\subsection{Long Term Credit Assignment in Markov Decision Processes via LRP and LSTMs}
\label{Arr:sec:Experiments_with_adapted_LSTMs}
% ----------------------------------------------------------------------------------------------------------------------------------------------
\index{Credit Assignment}
\index{Applications!Reinforcement Learning}
Assigning the credit for a received reward to actions that were performed
is one of the central tasks in reinforcement learning \cite{Arr:Sutton:17book}.
Long term credit assignment has been identified as one of the
biggest challenges in reinforcement learning \cite{Arr:Sahni:18}.
Classical reinforcement learning methods use a forward view approach by
estimating the future expected return of a Markov Decision Process (MDP).
However, they fail when the reward is delayed since they have 
to average over a large number of probabilistic future state-action paths
that increases exponentially with the delay of the reward \cite{Arr:Rahmandad:09,Arr:Luoma:17}.

In contrast to using a forward view, 
a backward view approach based on a backward analysis
of a forward model
avoids problems with unknown future state-action paths, 
since the sequence is already completed and known. 
Backward analysis transforms the forward view approach 
into a regression task, at which deep learning methods excel. 
As a forward model, an LSTM can be trained to predict the 
final return, given a sequence of state-actions.
LSTM was already used in reinforcement learning \cite{Arr:Schmidhuber:15}
for advantage learning \cite{Arr:Bakker:02} and
learning policies \cite{Arr:Hausknecht:15,Arr:Mnih:16,Arr:Heess:16}.
However, backward analysis via sensitivity analysis like ``backpropagation through a model''
\cite{Arr:Munro:87,Arr:Robinson:89,Arr:RobinsonFallside:89,Arr:Bakker:07} have
major drawbacks:
local minima, instabilities, exploding or vanishing
gradients in the world model, proper exploration,
actions being only regarded by sensitivity but not their contribution 
(relevance) \cite{Arr:Hochreiter:90,Arr:Schmidhuber:90diff}.

Contribution analysis, however, 
can be used to decompose the return prediction (the output relevance) into 
contributions of single state-action pairs along the observed sequence, 
obtaining a redistributed reward (the relevance redistribution).
As a result, 
a new MDP is created with the same optimal policies and, 
in the optimal case, with no delayed rewards (expected future rewards equal zero) \cite{Arr:Arjona-Medina:18}.
Indeed, for MDPs the $Q$-value is equal to the expected immediate reward 
plus the expected future rewards. Thus, if the expected future rewards are zero,
the $Q$-value estimation simplifies to computing the mean of the immediate
rewards.

In the following experiment we do not evaluate 
the performance of the agent under this reward redistribution.
Instead, the aim of this experiment is to show how
different LRP product rules
change the explanation of the model and, therefore,
the reward redistribution. 

\paragraph{LSTM and Markov Properties.}
% ----------------------------------------------------------------------------------------------------------------------------------------------
For LSTMs with forget gate or output gate \cite{Arr:Greff:15}, as described in Section~\ref{Arr:sec:LSTM}, the cell content does not comply to Markov assumptions.
This is because the current contribution of an input to a memory cell can be modified or hidden by later inputs, via the forget gate or output gate.
For example, the forget gate can erase or reduce the contribution of the current input in the future due to some later inputs.
Likewise, the output gate can hide the contribution of the current input by closing the gate until some future input opens the gate and reveals the already past contribution.

Figure~\ref{Arr:fig:3lstm}b shows an LSTM memory cell that
complies with the Markov property.

\paragraph{Environment.}
In our environment, an agent has to collect the {\em Moneybag} and then collect as many {\em Coins} as possible in a one dimension grid world.
Only {\em Coins} collected after collecting the {\em Moneybag} give reward.
At each time step, the agent can move to the left or to the right.
All rewards are only given at the end of the episode,
depending on how many {\em Coins} the agent collected in the {\em Moneybag}.

\paragraph{Training the Model.}
We are given a collection of sequences of state-action pairs. Each sequence represents one episode. Each episode is labeled with a scalar corresponding to the episode return. States and actions in the sequence are encoded as a vector of four binary features representing if the {\em Moneybag} is collected, if a {\em Coin} is collected, and the chosen action for the current timestep (one binary feature per action). 
In this experiment, we use a Long Short-Term Memory (LSTM) \cite{Arr:Hochreiter:91,Arr:Hochreiter:97} 
to predict the return of an episode \cite{Arr:Arjona-Medina:18}. Notice that since we are using an LSTM, the input encoding does not have to fulfil the Markov property. Once the LSTM is trained, we use LRP \cite{Arr:Bach:15} as contribution analysis for backward analysis.  

\paragraph{LRP and Different Product Rules.}
% ----------------------------------------------------------------------------------------------------------------------------------------------
\index{Evaluating Explanations!Indirect Evaluation}

\begin{figure}[!h]
\centering
\includegraphics[angle=0,width=0.80\textwidth]{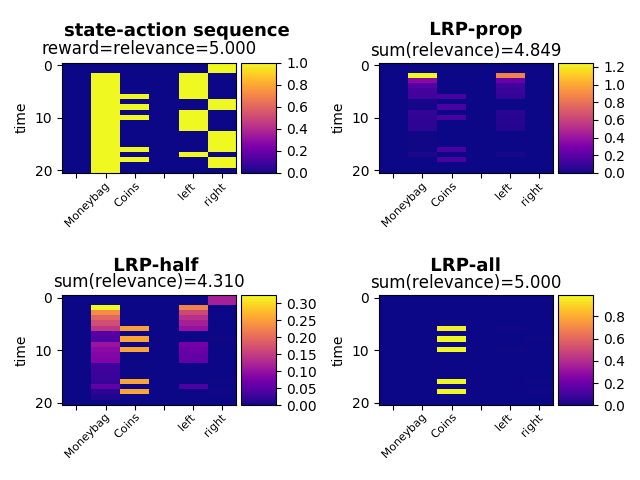}
\caption{LRP with different product rules as contribution analysis method in backward analysis, for one specific sequence of state-actions.
State and action sequence is represented as a vector of four binary features.
In this environment, {\em Coins} only give reward once the {\em Moneybag} is collected. When relevance is allowed to flow through the gate (LRP-prop and LRP-half rule), the event {\em Moneybag} is detected. Otherwise, only coin events are detected.
\label{Arr:fig:lrp_5coints}}
\end{figure}

\begin{figure}[!h]
\centering
\includegraphics[angle=0,width=0.80\textwidth]{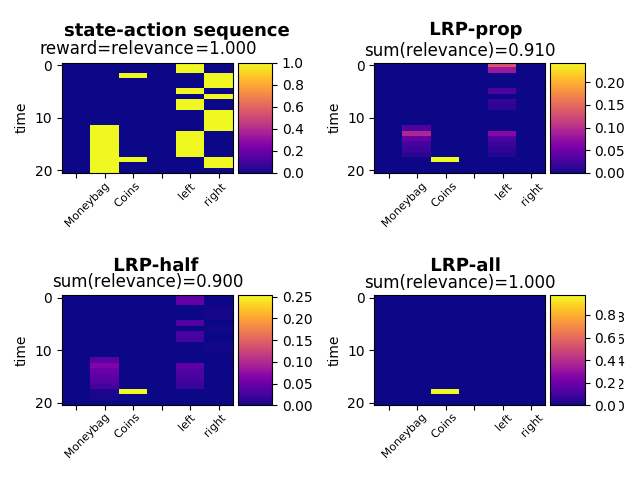}
\caption{LRP with different product rules as contribution analysis method in backward analysis, for one specific sequence of state-actions.
Since {\em Coins} without {\em Moneybag} do not count for the reward, no relevance is assigned to these events. 
\label{Arr:fig:lrp_coinbefore}}
\end{figure}

We trained an LSTM network, as depicted in Figure~\ref{Arr:fig:3lstm}b, to predict the return given a sequence of states and actions. Later, we applied different product rules to propagate the relevance through the input gates and the cell inputs (cf. Table~\ref{Arr:tab:product_rule_overview} Section~\ref{Arr:sec:LRP_gated_interactions}). Results are shown in Figure~\ref{Arr:fig:lrp_5coints} and~\ref{Arr:fig:lrp_coinbefore}.
When no relevance is propagated through the input gates (LRP-all),
certain important events are not detected by the contribution analysis,
i.e. no relevance is assigned to the event {\em Moneybag}.
This is due to the representation learned by the LSTM model,
which stores information about the {\em Moneybag} feature,
which in turn is used to activate a learned {\em Coins} counter via the input gate once the {\em Moneybag} has been collected.
As such, the {\em Moneybag} event contributes through the input gate.
When relevance is allowed to flow through the input gates (LRP-prop and LRP-half),
all events can be detected,
including the actions that lead to the {\em Moneybag} event.
However, the amount of relevance is not completely conserved when it is propagated through the gates,
as a small amount of relevance can get trapped in the LSTM cell, in particular via the relevance of the initial time step input gate.

\FloatBarrier

\section{Conclusion}
% ----------------------------------------------------------------------------------------------------------------------------------------------
We presented several ways of extending the LRP technique to recurrent neural networks such as LSTMs, which encompasses defining a rule to propagate the relevance through product layers.
Among the tested product rule variants we showed that, on standard LSTM models, the signal-take-all strategy leads to the most pertinent results,
and can be embedded in the deep Taylor framework where it corresponds to choosing the nearest root point in the gated interaction relevance model.

Additionally, we showed that the relevance propagation flow can be made more straightforward and stable by adapting the LSTM model towards the LRP technique
and that, in this case, propagating a share of relevance through the gates leads to a detection of relevant events earlier in time.
The resulting simplified and less connected LSTM model can potentially solve the same problems as the standard LSTM, although in practice this may necessitate using more memory cells.

More generally, further investigating the representational power of the new proposed LSTM, as well as its interplay with various LRP propagation rules, in particular via controlled experiments, would be a subject for future work.\\\\
{\bf Acknowledgements.} This work was supported by the German Ministry for Education and Research as Berlin Big Data Centre (01IS14013A), Berlin Center for Machine Learning (01IS18037I) and TraMeExCo (01IS18056A). Partial funding by DFG is acknowledged (EXC 2046/1, project-ID: 390685689). This work was also supported by the Institute for Information \& Communications Technology Planning \& Evaluation (IITP) grant funded by the Korea government (No.\ 2017-0-00451, No.\ 2017-0-01779).

\bibliographystyle{splncs04}

\end{document}